%% file: main.tex
\documentclass[sigconf,natbib=true]{acmart}

\setcopyright{none}                  % no copyright block in anonymous submission
\renewcommand\footnotetextcopyrightpermission[1]{}
\acmConference[CIKM '26]{Proceedings of the 35th ACM International
  Conference on Information and Knowledge Management}{November 7--11,
  2026}{Rome, Italy}
\acmYear{2026}
\acmISBN{}
\acmDOI{}

\usepackage{booktabs}
\usepackage{amsmath}
\usepackage{graphicx}
\usepackage{tikz}
\usetikzlibrary{positioning, arrows.meta, shapes.geometric,
                fit, backgrounds, calc}

\begin{document}

\title{Universal Pathologies, Conditional Consequences:
  A Triple-Robustness Analysis of RAG for Multi-Hop Traceability}

%% ArXiv preprint author block. For CIKM 2026 double-blind submission,
%% re-enable `anonymous=true' in the \documentclass options above and
%% revert this block to "Anonymous Author(s)".
\author{Meftun Akarsu}
\affiliation{%
  \institution{Technische Hochschule Ingolstadt}
  \city{Ingolstadt}
  \country{Germany}
}
\email{mea5963@thi.de}

\author{Burak \"Ozdemir}
\affiliation{%
  \institution{Independent Researcher}
  \city{Ankara}
  \country{Turkey}
}
\email{bozdemir@gmail.com}

\input{sections/01-abstract}

%% --- ACM CCS concepts --------------------------------------------------



\maketitle

\input{sections/02-introduction}
\input{sections/03-related-work}
\input{sections/04-method}
\input{sections/05-experiments}
\input{sections/06-results}
\input{sections/07-discussion}
\input{sections/08-conclusion}

%% --- GenAI disclosure (does NOT count toward 4-page limit) -------------
\input{sections/99-genai-disclosure}

%% --- References (do NOT count toward 4-page limit) ---------------------
\bibliographystyle{ACM-Reference-Format}
\bibliography{refs}

\end{document}

%% file: sections/01-abstract.tex
\begin{abstract}
GraphRAG underperforms vector RAG on citation precision in many reports,
but \emph{where} and \emph{why} have remained corpus-bound. We present a
triple-robustness analysis that holds the retrieval architecture fixed and
varies three orthogonal axes --- \textbf{embedder} (local e5-small $\to$
Azure text-embedding-3-small), \textbf{corpus} (DO-178C typed-edge
requirements $\to$ Wikipedia paragraph chains via MuSiQue),
and \textbf{judge} (paired GPT-5.4 $\times$ GPT-4.1) --- across 4{,}440
main-matrix runs, 600 cross-corpus runs, and 1{,}200 paired faithfulness
judgments. \textbf{(C2a)} Over-citation is architecturally universal:
GraphRAG emits 11--15 IDs per answer at citation precision 0.12--0.23 and
retrieval recall 0.68--0.87 across all three settings.
\textbf{(C2b)} Its faithfulness consequence is corpus-conditional: in
typed-edge DO-178C, GraphRAG faithfulness collapses 74\%$\to$40\% across
hops; on Wikipedia chains the same pipeline rises 42\%$\to$58\% because
over-cited paragraphs remain topically supporting.
\textbf{(C1)} Stratum-conditional winners are corpus-conditional but
embedder-robust: vanilla wins 2-hop on DO-178C, GraphRAG wins 2-hop on
MuSiQue, identical under either embedder.
\textbf{(C3)} Single-judge LLM faithfulness is fragile to retrieval state:
same-judge self-$\kappa$ across embedders is $0.137$ for GPT-5.4 (verdict
change on 41\% of items). A learned router on dense embeddings alone
reaches macro-$F_1$ $0.86$ on hop classification \textbf{(C4)}. We argue
triple-robustness is the minimum bar for trustworthy RAG architecture
claims.
\end{abstract}

%% file: sections/02-introduction.tex
\section{Introduction}
\label{sec:intro}

DO-178C-style aerospace requirements are authored as typed link graphs:
each requirement \textit{derives\_from} parents, \textit{satisfies}
system-level intent, and \textit{traces\_to} verification artifacts.
Certification authorities increasingly require auditable chains spanning
two or more hops~\cite{do178c,arp4754a,easa2025npa,faa2024airoadmap}.
RAG architectures for this task have produced contradictory verdicts:
Microsoft GraphRAG~\cite{graphrag2024} exploits structure but
under-performs vector RAG on detailed queries~\cite{ragvsgraphrag2025,
graphragbench2026}; Adaptive-RAG~\cite{adaptiverag2024} routes by
complexity yet is hop-blind; LiSSA~\cite{lissa2025} and
TVR~\cite{niu_tvr2025} omit typed-graph reasoning. The literature
describes \emph{what} sometimes works in \emph{which} setting; what is
missing is a mechanism-level account of \emph{why} the verdicts shift.

We address this with a \textbf{triple-robustness} analysis. Holding a
five-pipeline architecture matrix fixed (vanilla, agentic, agentic+graph,
GraphRAG, learned adaptive), we vary three orthogonal axes: the retrieval
embedder (e5-small 384d $\to$ Azure 3-small 1536d), the corpus (DO-178C
$\to$ Wikipedia paragraph chains via MuSiQue mapped to 1/2/3+-hop
strata), and the faithfulness judge (paired GPT-5.4 $+$ GPT-4.1; single
GPT-5.4 on MuSiQue). The design produces 4{,}440 main-matrix runs, 600
cross-corpus runs, and 1{,}200 paired judgments.

We make four contributions.
\textbf{(C2a)} Over-citation is an \emph{architectural universal}:
GraphRAG cites 11--15 IDs at citation precision 0.12--0.23 with retrieval
recall 0.68--0.87 across all three settings (Table~\ref{tab:pathology}).
\textbf{(C2b)} Its faithfulness consequence is \emph{corpus-conditional}:
DO-178C collapses 74\%$\to$40\%, MuSiQue rises 42\%$\to$58\%.
\textbf{(C1)} Stratum-conditional winners are
\emph{corpus-conditional but embedder-robust} (Table~\ref{tab:dominance}):
vanilla wins 2-hop on DO-178C; GraphRAG wins 2-hop on MuSiQue.
\textbf{(C3)} \emph{Single-judge faithfulness is embedder-fragile}: GPT-5.4
self-$\kappa$ across embedders is just $0.137$ on the paired 300-tuple
subset (Table~\ref{tab:judges}).
\textbf{(C4, supporting)} a logistic-regression router on dense embeddings
alone reaches macro-$F_1$ $0.86$ on hop classification.

\S\ref{sec:related} surveys related work; \S\ref{sec:method} the method;
\S\ref{sec:experiments} the protocol; \S\ref{sec:results} results;
\S\ref{sec:discussion} threats.

%% file: sections/03-related-work.tex
\section{Related Work}
\label{sec:related}

\noindent\textbf{Agentic and Adaptive RAG.}
Self-RAG~\cite{selfrag2024} and CRAG~\cite{crag2024} introduced reflective
and corrective retrieval. Adaptive-RAG~\cite{adaptiverag2024} routes by
predicted query complexity; Probing-RAG~\cite{probingrag2025} extends with
internal-state probes; Search-R1~\cite{searchr1_2025} trains agentic
retrieval via RL; RAG-Critic~\cite{ragcritic2025} adds an iterative critic.
None condition routing on typed-graph hop distance or expose a typed
graph-lookup tool.

\noindent\textbf{GraphRAG and multi-hop QA.}
Microsoft GraphRAG~\cite{graphrag2024} and follow-ups~\cite{lightrag2024,
hipporag2_2025,graphr1_2025} construct entity-relation graphs at indexing
time. Han et~al.~\cite{ragvsgraphrag2025} and GraphRAG-Bench~\cite{graphragbench2026}
report that GraphRAG \emph{does not} dominate vector RAG across all query
types --- our C2a confirms this with quantified over-citation metrics in
safety-critical RE, and C2b localises \emph{why} (mechanism vs.~consequence)
across corpora. MuSiQue~\cite{musique2022} and
MultiHop-RAG~\cite{multihoprag2024} are adjacent multi-hop benchmarks; we
use MuSiQue for cross-corpus replication of C2a only.

\noindent\textbf{RAG for requirements traceability.}
LiSSA~\cite{lissa2025}, TVR~\cite{niu_tvr2025}, and Graph-RAG for
compliance~\cite{graphragcompliance2024} evaluate single regulated domains
without hop stratification or dual-judge protocols. Our triple-robustness
design is the natural completion of this thread.

\noindent\textbf{Citation evaluation, judge bias, and the $\kappa$ paradox.}
ALCE~\cite{alce2023} introduced citation $P/R/F_1$ for grounded generation;
Wallat~et~al.~\cite{wallat2024} show that ALCE-style correctness is not
faithfulness, motivating LLM judges. RAGChecker~\cite{ragchecker2024}
provides a single-judge framework. Self-preference
bias~\cite{selfpref_bias2024} motivates judge ensembles, and the
Feinstein--Cicchetti $\kappa$ paradox~\cite{feinstein1990} with Gwet's
AC1~\cite{gwet2008} establishes the prevalence correction we apply. Our C3
extends this with same-judge self-$\kappa$ across embedders as a sharper
fragility statistic.

\noindent\textbf{Statistical protocol.}
Berg-Kirkpatrick~et~al.~\cite{bergkirkpatrick2012} and
Koehn~\cite{koehn2004} popularised paired permutation tests; we use BCa
bootstrap CIs with Holm-corrected Wilcoxon and Cliff's $\delta$.

%% file: sections/04-method.tex
\section{Method}
\label{sec:method}

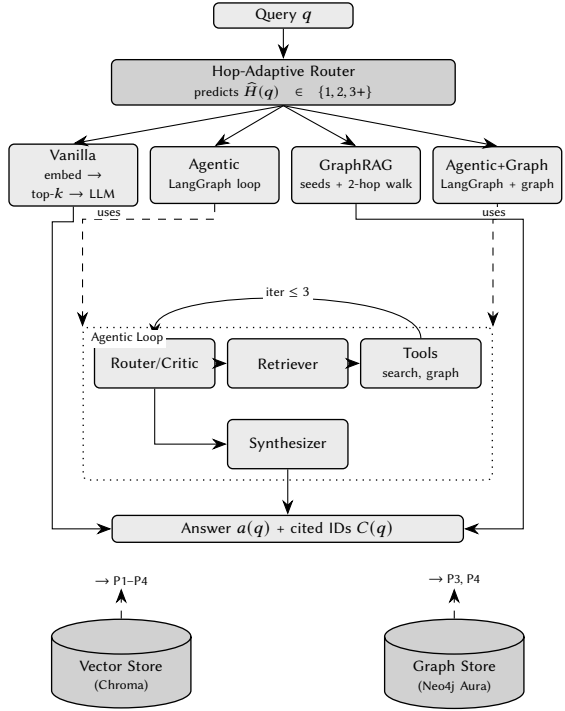
\begin{figure}[t!]
  \centering
  \input{figures/architecture-tikz}%
  \caption{Five-pipeline architecture with shared embedder, vector store,
    and typed-edge graph. Vanilla and GraphRAG bypass the agentic loop;
    Agentic and Agentic+Graph route through router/critic--retriever--tools
    (iter${\leq}3$) before synthesis. The adaptive pipeline (V1 rule-based
    or V2 learned) selects one of the four base pipelines per query.}
  \Description{Block diagram of the five-pipeline RAG architecture: a
    query enters a hop-adaptive router that selects one of four base
    pipelines (Vanilla, Agentic, GraphRAG, Agentic+Graph). The Agentic
    and Agentic+Graph pipelines invoke an inner Agentic Loop comprising
    a router/critic, retriever, tools (search and graph), and a
    synthesizer that iterates up to three times. Outputs combine into
    an answer with cited IDs. A shared vector store (Chroma) feeds
    P1--P4 and a shared graph store (Neo4j Aura) feeds P3 and P4.}
  \label{fig:architecture}
\end{figure}

\subsection{Pipelines}
\label{sec:pipelines}
All pipelines share embedder, ChromaDB vector store, Neo4j typed-edge
graph, Azure GPT-5.4 generator, and grounded synthesis prompt; they
differ only in retrieval. \textbf{(i)~Vanilla}: dense top-$k$ + one
synthesis call. \textbf{(ii)~Agentic}: LangGraph
router-retriever-critic loop, \texttt{search\_documents} tool only, iter
cap 3. \textbf{(iii)~GraphRAG}: vector seeds + up-to-2-hop typed-graph
walk via Cypher, single synthesis. \textbf{(iv)~Agentic+Graph}: (ii) plus
a typed-edge \texttt{graph\_lookup} tool. \textbf{(v)~Adaptive}: chooses
among (i)--(iv) per query; we evaluate a rule-based router (V1) and a
learned out-of-fold logistic regression (V2).

\subsection{Triple-Robustness Axes}
\label{sec:axes}
\textbf{Embedder:} local intfloat/multilingual-e5-small (384d) vs.\ Azure
\texttt{text-embedding-3-small} (1536d); ChromaDB collections rebuilt per
embedder, Neo4j shared. \textbf{Corpus:} a 1{,}132-requirement
DO-178C-style aerospace certification corpus across 32 modules,
LLM-anonymized from proprietary source artifacts for release
(\S\ref{sec:disclosure}); and a 200-query
MuSiQue~\cite{musique2022} subset (67/67/66 across 2/3/4-hop, mapped to
our 1/2/3+-hop strata, \textit{REFERENCES} edges only between consecutive
supporting paragraphs). \textbf{Judge:} GPT-5.4 $+$ GPT-4.1 on the
DO-178C main matrix; single-judge GPT-5.4 on MuSiQue.

\subsection{Metrics and Judging}
\label{sec:metrics}
We report ALCE-style citation $P/R/F_1$~\cite{alce2023} against gold
requirement-ID sets and retrieval recall on retrieved contexts. Each
faithfulness judgment is a strict-JSON binary verdict over the
retrieved-context block. We report Cohen's $\kappa$, Gwet's
AC1~\cite{gwet2008,feinstein1990}, raw agreement, McNemar exact-binomial
$p$, and Spearman $\rho$ on per-pipeline ranking; we further report
same-judge self-$\kappa$ across embedders on a paired 300-tuple subset
(\S\ref{sec:c3}).

\subsection{V2 Learned Router}
\label{sec:routerdesign}
14 features per query: 11 hand text features (ID regex, keyword flags,
log token length) and 3 PCs of the query embedding (fit per fold).
Multinomial logistic regression, $L_2$ ($C{=}1$), Platt-calibrated,
under stratified 10-fold $\times$ 5-repeat CV (50 fits). The per-stratum
routing target is the empirical mean-$F_1$ winner from the locked main
matrix; we report out-of-fold predictions for all 296 queries.

%% file: figures/architecture-tikz.tex
%% architecture-tikz.tex --- the tikzpicture body for Figure 1.
%% Single source of truth. Consumed in two ways:
%%   (a) figures/architecture.tex (standalone) for separate compile.
%%   (b) sections/04-method.tex via \input{figures/architecture-tikz}
%%       so the figure renders inline when main.tex is compiled.
%%
%% Required preamble in the host document:
%%   \usepackage{tikz}
%%   \usetikzlibrary{positioning, arrows.meta, shapes.geometric,
%%                   fit, backgrounds, calc}

\begin{tikzpicture}[
  node distance=4mm,
  >={Stealth[length=2mm]},
  every node/.style={font=\sffamily\scriptsize}
]
\tikzset{
  box/.style    ={draw=black, line width=0.5pt, rounded corners=2pt,
                  align=center, inner sep=2pt},
  light/.style  ={box, fill=black!8},
  mid/.style    ={box, fill=black!18},
  pipe/.style   ={light, text width=1.55cm, minimum width=1.7cm,
                  minimum height=7.5mm},
  lp/.style     ={light, text width=1.45cm, minimum width=1.55cm,
                  minimum height=6.5mm},
  store/.style  ={cylinder, shape border rotate=90, aspect=0.25,
                  draw=black, line width=0.5pt, fill=black!18,
                  align=center, inner sep=2pt,
                  text width=1.7cm, minimum height=7mm},
  arr/.style    ={->, line width=0.4pt},
  darr/.style   ={arr, dashed},
  lab/.style    ={font=\sffamily\tiny, fill=white, inner sep=1pt},
  group/.style  ={draw=black, line width=0.5pt, rounded corners=2pt,
                  dotted, inner sep=4pt}
}

%% ----- Top: Query, Router -----
\node[light, text width=1.7cm] (q) at (0,0) {Query $q$};
\node[mid, below=4mm of q, text width=4.4cm] (router)
  {Hop-Adaptive Router\\{\tiny predicts $\widehat{H}(q) \in \{1, 2, 3{+}\}$}};
\draw[arr] (q) -- (router);

%% ----- Pipelines (centered at x=0) -----
\node[pipe, below=5mm of router, xshift=-2.775cm] (p1)
  {Vanilla\\{\tiny embed $\to$ top-$k$ $\to$ LLM}};
\node[pipe, right=1.5mm of p1] (p2)
  {Agentic\\{\tiny LangGraph loop}};
\node[pipe, right=1.5mm of p2] (p3)
  {GraphRAG\\{\tiny seeds + 2-hop walk}};
\node[pipe, right=1.5mm of p3] (p4)
  {Agentic+Graph\\{\tiny LangGraph + graph}};
\foreach \i in {1,2,3,4}
  \draw[arr] (router.south) -- (p\i.north);

%% ----- Agentic Loop subgraph (centered below pipelines) -----
\node[lp] (critic) at (-1.7,-4.6) {Router/Critic};
\node[lp, right=1.5mm of critic] (retr)  {Retriever};
\node[lp, right=1.5mm of retr]   (tools) {Tools\\{\tiny search, graph}};
\node[lp, below=4mm of retr]     (synth) {Synthesizer};

\draw[arr] (critic) -- (retr);
\draw[arr] (retr)   -- (tools);
\draw[arr] (tools.north) to[out=90,in=90,looseness=0.5]
  node[lab, midway, yshift=1mm] {iter $\le 3$} (critic.north);
\draw[arr] (critic.south) |- (synth.west);

\begin{scope}[on background layer]
  \node[group, fit=(critic)(retr)(tools)(synth)] (loop) {};
\end{scope}
\node[lab, anchor=north west]
  at ([xshift=2pt,yshift=-1pt]loop.north west) {Agentic Loop};

%% ----- P2, P4 -> Loop (dashed "uses"; right-angle elbows) -----
\draw[darr] (p2.south) -- ++(0,-2mm) -| (loop.north west)
  node[lab, pos=0.4, above] {uses};
\draw[darr] (p4.south) -- ++(0,-2mm) -| (loop.north east)
  node[lab, pos=0.4, above] {uses};

%% ----- Output -----
\node[light, below=6mm of synth, text width=4.5cm] (out)
  {Answer $a(q)$ + cited IDs $C(q)$};
\draw[arr] (synth.south) -- (out.north);

%% ----- P1, P3 -> Output (bypass loop on the sides) -----
\draw[arr] (p1.south) -- ++(0,-2mm) -|
  ([xshift=-4mm]loop.south west) |- (out.west);
\draw[arr] (p3.south) -- ++(0,-2mm) -|
  ([xshift=4mm]loop.south east)  |- (out.east);

%% ----- Data stores at the bottom; consolidated arrows -----
\node[store, below=10mm of out, xshift=-22mm] (s1)
  {Vector Store\\{\tiny (Chroma)}};
\node[store, below=10mm of out, xshift=22mm]  (s2)
  {Graph Store\\{\tiny (Neo4j Aura)}};

\draw[darr] (s1.north) -- ++(0,4mm)
  node[lab, anchor=south] {$\to$ P1--P4};
\draw[darr] (s2.north) -- ++(0,4mm)
  node[lab, anchor=south] {$\to$ P3, P4};

\end{tikzpicture}

%% file: sections/05-experiments.tex
\section{Experimental Setup}
\label{sec:experiments}

\textbf{Main matrix and cross-corpus sample.} DO-178C: 5 pipelines $\times$
296 hop-stratified queries $\times$ 3 seeds $= 4{,}440$ runs, executed
under both embedders (\emph{v2}, \emph{v3}). MuSiQue: 3 pipelines (vanilla,
GraphRAG, agentic+graph) $\times$ 200 queries $\times$ 1 seed $= 600$ runs,
Azure embedder only. Agentic and adaptive are not re-run on MuSiQue (C2a
needs only the three-pipeline trio).

\textbf{Faithfulness judging.} On DO-178C, multi-judge protocol
(GPT-5.4 $+$ GPT-4.1) is applied to a 300-row stratified subset (60 per
pipeline, seed 42), yielding 600 paired binary judgments per embedder.
The v3 judged subset is explicitly \emph{pinned} to the same 300
(query, pipeline, repeat) tuples as v2 so C3 deltas are computed on
identical items. On MuSiQue, we single-judge all 600 with GPT-5.4.

\textbf{Statistical protocol.} 95\% BCa bootstrap intervals~\cite{du2025bootstrap}
($B{=}1000$, paired at the query level), Wilcoxon signed-rank with Holm
correction across the 30 pipeline-pair $\times$ stratum contrasts, and
Cliff's $\delta$ (negligible ${<}0.147$; Macbeth 2011). A pipeline pair
is reported as significantly different only when the BCa interval excludes
zero, Holm $p{<}0.05$, and $|\delta|\!\geq\!0.147$ jointly hold.

\textbf{Reproducibility.} The full pipeline regenerates from one
\texttt{Makefile} target on locked CSVs; seeds, sampling protocols, and
Azure model snapshots are in supplementary materials. The MuSiQue
subgraph (3{,}996 paragraph chunks, 399 \textit{REFERENCES} edges) is
built deterministically from \texttt{dgslibisey/MuSiQue} (validation
split, seed $20260511$).

%% file: sections/06-results.tex
\section{Results}
\label{sec:results}

\input{tables/tab_dominance}

\begin{figure}[t!]
  \centering
  \includegraphics[width=\columnwidth]{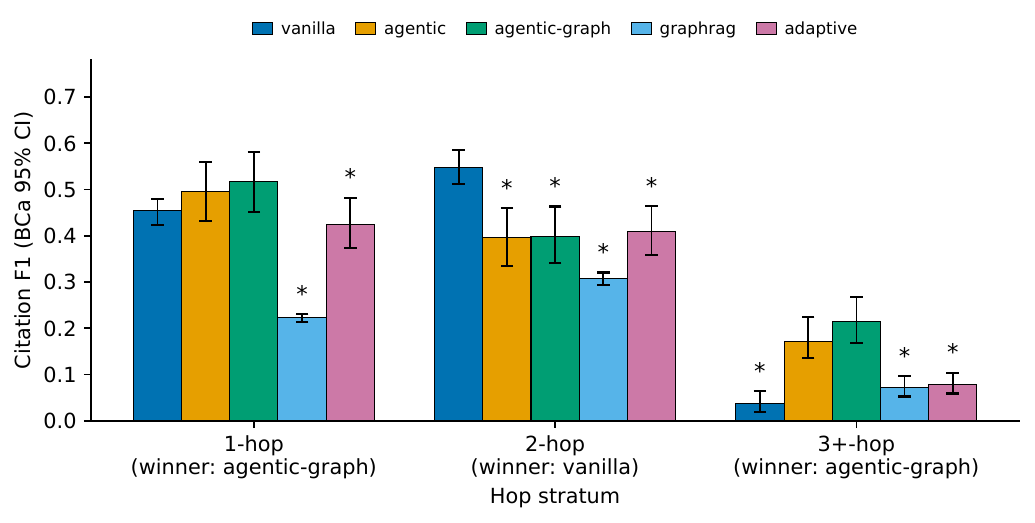}
  \caption{Per-stratum citation $F_1$ on the v2 main matrix (DO-178C,
    local e5-small embedder). Bars: the five pipelines across the
    1-/2-/3+-hop strata with BCa 95\% CIs; \mbox{$*$} marks pipelines
    significantly different from the per-stratum winner (Holm-Wilcoxon
    $p{<}0.05$, $|\delta|{\geq}0.147$). Cross-corpus and cross-embedder
    comparisons are in Table~\ref{tab:dominance}.}
  \Description{Grouped bar chart of per-stratum citation F1 for five
    RAG pipelines (vanilla, agentic, agentic-graph, graphrag, adaptive)
    on the DO-178C v2 main matrix, broken down by 1-hop, 2-hop, and
    3+-hop strata, with error bars and asterisks marking pipelines
    significantly different from the per-stratum winner.}
  \label{fig:perstratum}
\end{figure}

\subsection{C1: Winners are Corpus-Conditional, Embedder-Robust}
\label{sec:c1}
Table~\ref{tab:dominance} reports per-stratum $F_1$ across three settings.
On DO-178C the dominance pattern is identical under both embedders:
agentic-graph wins 1-hop and 3+-hop, vanilla wins 2-hop. The 1-hop and
3+-hop wins do not separate agentic-graph from agentic (Cliff's
$\delta{<}0.147$ in every stratum) and are read as an \{agentic-graph,
agentic\} tie. On MuSiQue the winners shift: agentic-graph retains 1-hop,
but GraphRAG wins both 2-hop ($F_1{=}0.375$) and 3+-hop ($F_1{=}0.342$),
reversing vanilla's 2-hop dominance. The pattern is corpus-conditional,
not universal.

\input{tables/tab_pathology}

\begin{figure}[t!]
  \centering
  \includegraphics[width=\columnwidth]{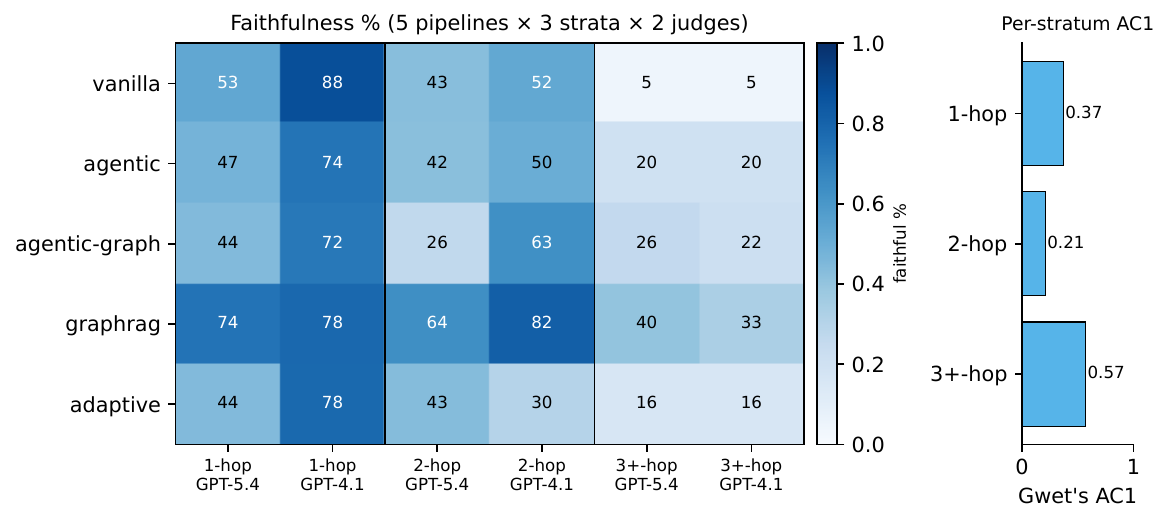}
  \caption{Multi-judge faithfulness on the v2 main matrix.
    \textit{Left}: faithfulness fraction (5 pipelines $\times$ 3 hop
    strata $\times$ 2 judges); the GraphRAG row exhibits the monotonic
    collapse ($74\%{\to}64\%{\to}40\%$ under GPT-5.4) that drives C2b
    on DO-178C. \textit{Right}: per-stratum Gwet AC1, surfacing the
    $\kappa$ paradox at the 3+-hop stratum (AC1${=}0.57$ at high
    prevalence; cf.\ Table~\ref{tab:judges}). v3 and MuSiQue panels
    are reported in Table~\ref{tab:pathology}.}
  \Description{Heatmap of faithfulness percentages for five RAG
    pipelines (rows: vanilla, agentic, agentic-graph, graphrag,
    adaptive) across three hop strata times two judges (columns:
    1-hop GPT-5.4, 1-hop GPT-4.1, 2-hop GPT-5.4, 2-hop GPT-4.1,
    3+-hop GPT-5.4, 3+-hop GPT-4.1) on the v2 main matrix. The
    graphrag row darkens monotonically left-to-right under both
    judges. A side panel shows per-stratum Gwet AC1 bars.}
  \label{fig:faithheatmap}
\end{figure}

\subsection{C2: GraphRAG's Universal Pathology, Conditional Consequence}
\label{sec:c2}
\noindent\textbf{(C2a) Over-citation is architectural.}
The top block of Table~\ref{tab:pathology} establishes universality.
Across embedder \emph{and} corpus swaps GraphRAG cites a mean of 11--15
IDs per answer at citation precision 0.12--0.23 with retrieval recall
in the 0.68--0.87 range. This is not a corpus-specific artifact --- the same
mechanism manifests on Wikipedia paragraph chains with a different gold
structure --- and converges with prior cross-domain
observations~\cite{ragvsgraphrag2025,graphragbench2026,wallat2024}.

\noindent\textbf{(C2b) The faithfulness consequence is corpus-conditional.}
The bottom block reveals the consequence split. On DO-178C with local
embedder, GraphRAG faithfulness collapses monotonically across hops
(74\%/64\%/40\%). Under embedder swap it flattens to ${\approx}50\%$. On
MuSiQue it \emph{rises} monotonically (42\%/54\%/58\%). The mechanistic
reading: on typed-edge regulated corpora, over-cited IDs are adversarial
distractors whose claims contradict the question; on Wikipedia paragraph
chains over-cited paragraphs remain topically related to the question
and typically satisfy the judge's support criterion. Over-citation is
therefore architecturally universal but consequentially domain-bound ---
a constraint on inferring faithfulness from citation precision alone.

\input{tables/tab_judges}

\subsection{C3: Single-Judge Faithfulness is Embedder-Fragile}
\label{sec:c3}
Table~\ref{tab:judges} reports inter-judge agreement on the paired
300-tuple subset and same-judge self-$\kappa$ across embedders.
Inter-judge $\kappa$ between GPT-5.4 and GPT-4.1 halves under embedder
swap on identical items (0.30 $\to$ 0.17 overall). Per-stratum Gwet AC1
surfaces the $\kappa$ paradox at the 3+-hop stratum (AC1 ${=}0.57$ in
v2, $0.35$ in v3, with $\kappa{<}0.10$ in both). The headline is the
same-judge row: \emph{the same judge reading the same question and
pipeline output changes verdict on 41\% of items when only the
retrieval embedding changes} ($\kappa_{\text{GPT-5.4}}{=}0.14$).
GPT-4.1 is materially more stable ($\kappa{=}0.48$). To our knowledge
this is the first quantification of single-judge fragility to upstream
retrieval state in a controlled paired design.

\input{tables/tab_router}

\paragraph{C4 (supporting): Dense embeddings alone classify hop.}
Table~\ref{tab:router} shows the V2 router replication. The 50-fold
macro-$F_1$ improves $0.78\pm0.08\to 0.86\pm0.05$ under embedder swap,
and Adaptive-V2 closes 56\% of the V1$\to$Oracle gap with Holm-significant
gains in all three strata under both embedders. Of 11 hand text features
only four (\texttt{kw\_which}, \texttt{kw\_satisfies},
\texttt{kw\_references}, \texttt{kw\_verifies}) are ever non-zero on the
296 queries; the rest are inert. The router therefore operates almost
exclusively on the 3 query-embedding PCs, suggesting hop distance is
largely an embedding-decodable property of the query text in
regulated-domain corpora.

%% file: tables/tab_dominance.tex
\begin{table*}[t]
  \centering
  \caption{Per-stratum citation $F_1$ across three settings. v2 main = DO-178C corpus with local e5-small embeddings (4{,}440 runs); v3 main = same corpus with Azure text-embedding-3-small (4{,}440 runs); MuSiQue = 200-query Wikipedia stratified subset with Azure embedder (600 runs, vanilla / agentic-graph / graphrag only). Bold marks the per-(setting, stratum) maximum.}
  \label{tab:dominance}
  \small
  \begin{tabular}{l ccc | ccc | ccc}
    \toprule
     & \multicolumn{3}{c|}{v2 main (local)} & \multicolumn{3}{c|}{v3 main (Azure)} & \multicolumn{3}{c}{MuSiQue (Azure)} \\
    System & 1h & 2h & 3+h & 1h & 2h & 3+h & 1h & 2h & 3+h \\
    \midrule
    vanilla & 0.454 & \textbf{0.548} & 0.037 & 0.486 & \textbf{0.558} & 0.111 & 0.439 & 0.340 & 0.256 \\
    agentic & 0.497 & 0.396 & 0.172 & 0.490 & 0.408 & 0.201 & --- & --- & --- \\
    agentic-graph & \textbf{0.517} & 0.398 & \textbf{0.215} & \textbf{0.500} & 0.402 & \textbf{0.249} & \textbf{0.457} & 0.296 & 0.249 \\
    graphrag & 0.224 & 0.308 & 0.071 & 0.230 & 0.306 & 0.114 & 0.358 & \textbf{0.375} & \textbf{0.342} \\
    adaptive & 0.425 & 0.410 & 0.079 & 0.449 & 0.402 & 0.119 & --- & --- & --- \\
    \bottomrule
  \end{tabular}
\end{table*}

%% file: tables/tab_pathology.tex
\begin{table}[t]
  \centering
  \caption{GraphRAG triple-robustness: (top, C2a) over-citation mechanism is universal across embedder and corpus swaps; (bottom, C2b) faithfulness consequence is corpus-conditional --- collapses on DO-178C but stays flat/rises on MuSiQue.}
  \label{tab:pathology}
  \small
  \begin{tabular}{l ccc}
    \toprule
    \textbf{C2a: over-citation mechanism} & v2 main & v3 main & MuSiQue \\
    \midrule
    Mean IDs cited & 14.9 & 14.9 & 11.2 \\
    Citation precision & 0.120 & 0.129 & 0.227 \\
    Retrieval recall & 0.681 & 0.724 & 0.873 \\
    Citation $F_1$ overall & 0.202 & 0.217 & 0.358 \\
    \midrule
    \textbf{C2b: faithfulness consequence} & 1-hop & 2-hop & 3+-hop \\
    \midrule
    v2 main (GPT-5.4) & 0.74 & 0.64 & 0.40 \\
    v3 main pinned (GPT-5.4) & 0.52 & 0.55 & 0.40 \\
    MuSiQue (GPT-5.4) & 0.42 & 0.54 & 0.58 \\
    \bottomrule
  \end{tabular}
\end{table}

%% file: tables/tab_judges.tex
\begin{table}[t]
  \centering
  \caption{Multi-judge protocol fragility. Top: inter-judge agreement (GPT-5.4 vs GPT-4.1) drops by half overall under embedder swap (v2 local $\to$ v3 Azure) on the same 300 paired tuples. Bottom: same-judge self-consistency across embedders --- GPT-5.4 changes verdict on 41\% of items when only the retrieval embedding changes; GPT-4.1 is more stable.}
  \label{tab:judges}
  \small
  \begin{tabular}{l rrrr}
    \toprule
    \textbf{Inter-judge $\kappa$} & 1-hop & 2-hop & 3+-hop & overall \\
    \midrule
    v2 main & 0.28 & 0.22 & 0.04 & 0.30 \\
    v3 main (pinned) & 0.27 & 0.07 & 0.07 & 0.17 \\
    \midrule
    \textbf{Gwet's AC1} & 1-hop & 2-hop & 3+-hop & overall \\
    \midrule
    v2 main & 0.37 & 0.21 & 0.57 & 0.31 \\
    v3 main (pinned) & 0.22 & 0.07 & 0.35 & 0.17 \\
    \midrule
    \multicolumn{5}{l}{\textbf{Same-judge self-$\kappa$ across embedders (paired 300 tuples)}} \\
    \midrule
    GPT-5.4 (v2 main vs v3 main) & \multicolumn{4}{r}{$\kappa = 0.137$, raw agreement 0.59} \\
    GPT-4.1 (v2 main vs v3 main) & \multicolumn{4}{r}{$\kappa = 0.480$, raw agreement 0.74} \\
    \bottomrule
  \end{tabular}
\end{table}

%% file: tables/tab_router.tex
\begin{table}[t]
  \centering
  \caption{V2 router replication and Azure-embedder boost. Same 14-feature schema (11 hand text features, 3 PCs of the query embedding); only the embedder differs. Text features remain mostly inert (4/11 non-zero), so the +0.08 macro-$F_1$ gain is attributable to dense-embedding hop-decodability.}
  \label{tab:router}
  \small
  \setlength{\tabcolsep}{3pt}
  \resizebox{\columnwidth}{!}{%
  \begin{tabular}{l rr}
    \toprule
     & v2 (local 384d $\to$ PCA-3) & v3 (Azure 1536d $\to$ PCA-3) \\
    \midrule
    50-fold macro-$F_1$ & $0.78 \pm 0.08$ & $\mathbf{0.86 \pm 0.05}$ \\
    Adaptive-V2 overall $F_1$ & 0.419 & 0.443 \\
    Adaptive-V1 overall $F_1$ & 0.307 & 0.325 \\
    Oracle overall $F_1$ & 0.512 & 0.535 \\
    V2 $-$ V1 (absolute) & $+0.112$ & $+0.118$ \\
    Gap closure (V1$\to$Oracle) & 54.7\% & 56.2\% \\
    Holm-Wilcoxon strata sig & 3/3 & 3/3 \\
    \bottomrule
  \end{tabular}%
  }
\end{table}

%% file: sections/07-discussion.tex
\section{Discussion}
\label{sec:discussion}

\paragraph{Why does over-citation's consequence depend on corpus?}
GraphRAG's graph traversal expands candidates beyond what dense retrieval
surfaces; the synthesizer then dutifully cites every ID it touches. On
DO-178C, typed edges connect requirements with adversarial specificity:
\textit{derives\_from} expansion drags in unrelated parent requirements;
\textit{references} expansion pulls in cross-module artifacts whose
claims contradict the question --- the judge marks the resulting
citations unsupported. On Wikipedia paragraph chains the over-cited
content is by construction topically adjacent; the judge typically finds
support in the expanded context even when many cites are extraneous.
The same retrieval behaviour therefore has different faithfulness
consequences. We predict the same coupling in any regulated-domain RAG
that exposes typed-edge expansion over an adversarial gold structure.

\paragraph{Implications for LLM-judge practice.}
Same-judge self-$\kappa{=}0.137$ across embedders means: comparing
faithfulness across retrieval modules with different embedders, using a
single judge, reports embedder-conditional verdicts, not invariant
faithfulness. Multi-judge ensembles mitigate only partially: inter-judge
$\kappa$ itself halves under embedder swap. Triple-robustness, or at
minimum embedder-controlled paired-tuple judging, is the minimum
defensible bar for LLM-judge faithfulness.

\paragraph{Threats to validity.}
\textit{Construct.} ALCE-style $F_1$ does not measure rationale quality;
we pair it with multi-judge faithfulness, while acknowledging LLM-judge
bias~\cite{selfpref_bias2024,wallat2024}. C3 is the explicit reckoning.
\textit{External.} The DO-178C corpus is a single proprietary aerospace
certification dataset that has been LLM-anonymized for release because
the raw source artifacts are subject to IP and export-control
restrictions (\S\ref{sec:disclosure}); the absolute $F_1$ levels reported
here therefore characterize this single regulated-domain setting and may
not transfer to other proprietary or open-domain projects. The MuSiQue
cross-corpus replication (C2a) is the load-bearing external-validity
check for the architectural-universal claim; cross-corpus replication of
C1 and C2b is left for the journal extension.
\textit{Statistical.} Per-stratum $n\!\in\![95,111]$ is at the lower edge
of BCa stability under heavy skew; we report Cliff's $\delta$ alongside
intervals and observed no qualitative reversal under a percentile
fallback. \textit{Reranker / router engineering.} No cross-encoder
reranker is included in the main pipelines (orthogonal axis, future
work); C4 is framed as a diagnostic instrument revealing hop-class
decodability, not a recommended production routing system.

%% file: sections/08-conclusion.tex
\section{Conclusion}
\label{sec:conclusion}
We presented a triple-robustness analysis (embedder $\times$ corpus
$\times$ judge) of five RAG architectures for multi-hop requirements
traceability. The empirical picture is sharper than either uniform
endorsement or uniform critique of GraphRAG. \emph{Over-citation is
architecturally universal}; \emph{its faithfulness consequence is
corpus-conditional}; \emph{stratum-conditional winners are
corpus-conditional but embedder-robust}; and \emph{single-judge
faithfulness verdicts are embedder-fragile in a way multi-judge protocols
only partially mitigate}. We argue that triple-robustness is the minimum
defensible bar for RAG architecture claims and that a learned hop router
on dense embeddings alone is a useful diagnostic instrument for
characterising the structure of the problem.

%% file: sections/99-genai-disclosure.tex
\section*{GenAI Usage Disclosure}
\label{sec:disclosure}
We disclose all uses of generative AI in this work in accordance with
the ACM Policy on Authorship and the CIKM 2026 disclosure requirement.
\textbf{(a)~Corpus anonymization.} The DO-178C-style corpus used in
this paper is derived from proprietary aerospace certification
artifacts that cannot be released directly due to intellectual-property
and export-control restrictions. Azure OpenAI GPT-5.4 was used under
a structured reformatting prompt (provided in the supplementary
material) to anonymize and rewrite the source material into 1{,}132
standalone requirement records while preserving the original typed-edge
structure (\textit{derives\_from}, \textit{satisfies},
\textit{references}, \textit{traces\_to}, \textit{verifies}) and
module taxonomy. All resulting records were programmatically validated
for format and uniqueness; no semantic validation or fact synthesis
was performed by GenAI.
\textbf{(b)~Pipeline components.} GPT-5.4 is the generation backbone
of all five RAG pipelines under study; it is the \textit{object} of
evaluation, not an authorial tool.
\textbf{(c)~LLM-as-judge.} GPT-5.4 and GPT-4.1 are used as the two
faithfulness judges in the locked dual-judge protocol; this is part
of the paper's experimental analysis and is fully documented in
\S\ref{sec:experiments}.
\textbf{(d)~Writing.} GenAI tools (Claude Sonnet 4.5, GPT-5.4) were
used for prose polishing, LaTeX-table assembly, and code-comment
generation. No section of the paper was authored \textit{de novo} by
an LLM. All claims, numbers, citations, and conclusions were verified
by the author(s). The author(s) take full responsibility for the
content.